# Tightly Coupled Optimization-based GPS-Visual-Inertial Odometry with Online Calibration and Initialization

Shihao Han, Feiyang Deng, Tao Li, and Hailong Pei

*Abstract*—In this paper, we present a tightly coupled optimization-based GPS-Visual-Inertial odometry system to solve the trajectory drift of the visual-inertial odometry especially over long-term runs. Visual reprojection residuals, IMU residuals, and GPS measurement residuals are jointly minimized within a local bundle adjustment, in which we apply GPS measurements and IMU preintegration used for the IMU residuals to formulate a novel GPS residual. To improve the efficiency and robustness of the system, we propose a fast reference frames initialization method and an online calibration method for GPS-IMU extrinsic and time offset. In addition, we further test the performance and convergence of our online calibration method. Experimental results on EuRoC datasets show that our method consistently outperforms other tightly coupled and loosely coupled approaches. Meanwhile, this system has been validated on KAIST datasets, which proves that our system can work well in the case of visual or GPS failure.

## I. INTRODUCTION AND RELATED WORK

In recent years, various sensors like cameras, IMU and GPS have been applied in modern autonomous robot systems to achieve accurate and high-rate pose estimates [1]–[7], since these sensors are low-cost yet provide high-performance ego-motion measurements. However, separately using these sensors is difficult to get an accurate and drift-free estimation in long-term autonomous navigation. For example, the global positioning system can acquire geodetic coordinates in outdoor scenes, but the GPS measurements are noisy and inaccurate in urban canyons. In contrast, Visual Inertial Odometry can achieve accurate and robust pose estimates in urban canyons, but the trajectory of the visual-inertial odometry will suffer from drift in long-term autonomous navigation. Thus, to achieve more accurate and robust localization, sensor fusion has received considerable attention. As shown in Fig. 1, the GPS and Visual Inertial Odometry fusion system achieves more accurate results

To fuse GPS and VIO, the transformation between the reference frames of the GPS and the VIO is needed. We usually use the starting point as the datum to convert the GPS measurements from the geodetic coordinate system to the East-North-Up (ENU) coordinate system. Besides, the VIO system estimates its state relative to the local world frame. As the $z^w$ axis of the local world frame is aligned with the gravity direction after each VIO system initialization, the transformation between the reference frames of GPS and VIO will also change accordingly. Therefore, the transformation with online

This work was partially supported by XAG company.
Shihao Han, Tao Li, and Hailong Pei are with the School of Automation Science and Engineering, South China University of Technology, Guangzhou, China. (e-mail: {hanshihao.j, renmengqisheng}@gmail.com ).
Feiyang Deng is with the Department of Electrical Engineering, City University of Hong Kong, Hong Kong, China. (e-mail: feiyadeng2-c@my.cityu.edu.hk).

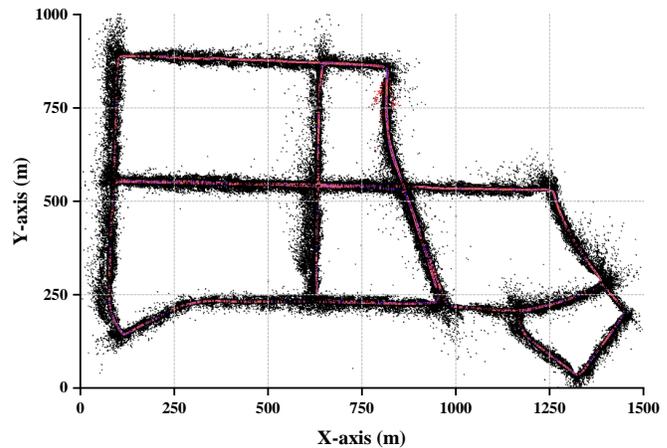

Fig. 1. The point cloud and trajectory of our tightly coupled system in the KAIST urban 28. The total length of the trajectory is 11.47km.

initialization is crucial in GPS and VIO's fusion process, because this transformation can align trajectories provided by GPS and VIO in different systems respectively. In [8], S. Umeyama presented a method to compute the transformation between two trajectories solving the problem of Online Initialization. [9] proposed a reference frame initialization way that leverages the solution to a least-squares problem with quadratic constraint. An online coarse-to-fine approach [10] initializes GNSS-visual-inertial states through the following three steps: Coarse Anchor Point Localization, Yaw Offset Calibration, and Anchor Point Refinement. In addition, one-shot alignment using a part of the sequence is not the best, since GPS measurements are noisy or the states of VIO may be wrong during the IMU initialization. Thus, the transformation process can be continuously optimized during the VIO system running.

Another important question during the fusion of GPS and VIO is the GPS-IMU extrinsic calibration. The extrinsic initial value is usually obtained via manual measurement but causing measurement errors, so we need an online GPS-IMU extrinsic calibration. [11]–[14] introduce researches about online calibrations of GPS, IMU, and camera within a Kalman Filter (KF) framework. [9] also proposed a method to perform online calibration of both the extrinsic and time offset within an Extended Kalman Filter (EKF) framework.

There are two main approaches to fuse GPS measurements and VIO states at the system level. The first approach is an EKF-based algorithm, MSCKF, which derives geometric constraints between multiple cameras' poses to optimize the system states efficiently. Based on MSCKF, [9] proposed a tightly coupled estimator to fuse GPS measurements, IMU, and camera. Besides, it is the first work using an online calibration approach for time offset between GPS and VIO. However, this approach generally produces slightly poor position precision compared with the nonlinear optimization approach.

Another approach is nonlinear optimization. In [10,15], a non-linear optimization-based visual-inertial odometry tightly coupled with raw GNSS pseudo-range and Doppler shift measurements has been presented, and they used raw GPS data rather than resolved GPS data. Paper [16] is a loosely coupled estimator that fuses GPS measurements and states of VIO in the optimization based on [17]. But this article did not support online calibration for GPS-IMU extrinsic and time offset. G. Cioffi used IMU preintegration to efficiently compute the global position residual [18]. However, this paper assumed the GPS and IMU data are synchronized, and online calibration for GPS-IMU extrinsic and time offset is also not supported. Furthermore, the paper [18] needs to compute new IMU preintegration and corresponding covariance, and new IMU preintegration must be continuously updated during the iterative optimization process, which increases the system's complexity.

In this paper, we propose an optimization-based approach that tightly couples camera, IMU, and GPS measurements based on [19], while focusing on online GPS-IMU extrinsic and time offset calibration. Besides, we present a novel fast initialization method of GPS-VIO that improves the efficiency of initialization. We highlight our contribution as follows:

- We formulate a novel GPS residual utilizing GPS measurements and IMU preintegration (same as the one used for the IMU residual). Since the GPS residual or visual can correct the IMU bias, our method can maintain the system's regular work in the case of visual or GPS failure.
- We propose an online calibration method for GPS-IMU extrinsic and time offset under a non-linear optimization framework, then we further test the performance and convergence of our online GPS-IMU extrinsic and time offset calibration method.
- We propose a fast initialization method of the transformation between the reference frames of GPS and VIO, which improves the system makes effective use of GPS.
- We evaluate our proposed method in comparison with another tightly coupled approach [18] on the EuRoC dataset by assuming the same magnitude of the noise, which proves that our system's positional accuracy is superior to [18]. Meanwhile, compared with other tightly coupled approaches, our system is the first to be tested on the common KAIST dataset. In addition, since there is no open sourced tightly coupled GPS-VIO system, we compare our system with an open sourced loose coupling approach VINS-Fusion on KAIST and our outdoor datasets.

I. PRELIMINARIES

A. Frames and Notations

The sensor frames and the world frames are depicted in Fig. 2, in which $(\cdot)^b$, $(\cdot)^c$ and $(\cdot)^g$ represent the IMU frame, camera frame, and GPS frame, respectively. $(\cdot)^{b_k}$, $(\cdot)^{c_k}$ and $(\cdot)^{g_k}$ respectively denote the parameters of the three frames at time $t_k$. And we use $(\cdot)^{enu}$ to denote the ENU frame and $(\cdot)^w$ to denote the local world frame, where both the ENU frame's $z^{enu}$

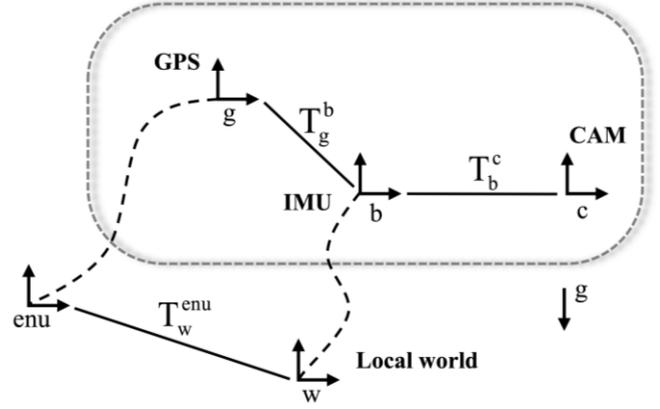

Fig. 2. An illustration of the sensor frames and the world frames.

axis and the local world frame's $z^w$ axis are parallel to the gravity direction.

For the world frames, $p^w_{b_i}$ and $R^w_{b_i}$ respectively represent the position and orientation of the IMU frame concerning the local world frame at time $t_i$. And $v^w_{b_i}$ is the velocity of the IMU frame measured in the local world frame at time $t_i$. Besides, $b_{a_i}$ is the accelerometer bias at time $t_i$, $b_{w_i}$ is the gyroscope bias at time $t_i$ and $p^{enu}_{g_i}$ represents the GPS measurement at time $t_i$. The position and orientation between the first IMU frame and ENU frame are represented as $p^{enu}_{b_0}$ and $R^{enu}_{b_0}$. We use $p^w_{b_0}$ and $R^w_{b_0}$ to represent the position and orientation of the first IMU frame concerning the local world frame. Finally, $p^{enu}_w$ and $R^{enu}_w$ represent the transformation between the reference frames of GPS and IMU.

B. Tightly Coupled Optimization-based GPS-VIO

According to ORB-SLAM3, the state vectors can be defined by the equation (1):

$$\mathcal{S}_i \doteq \{T^w_{b_i}, v^w_{b_i}, b_{a_i}, b_{w_i}\} \quad (1)$$

and given the time interval $[t_i, t_{i+1}]$, the definition of inertial residual is presented in equations (2):

$$\begin{cases} r_{I_{i,i+1}} = [r_{\Delta R_{i,i+1}}, r_{\Delta v_{i,i+1}}, r_{\Delta p_{i,i+1}}] \\ r_{\Delta R_{i,i+1}} = \log(\Delta R^T_{i,i+1} R^T_i R_{i+1}) \\ r_{\Delta v_{i,i+1}} = R^T_i(v_{i+1} - v_i - g\Delta t_{i,i+1}) - \Delta v_{i,i+1} \\ r_{\Delta p_{i,i+1}} = R^T_i\left(p_j - p_i - v_i \Delta t - \frac{1}{2}g\Delta t^2\right) - \Delta p_{i,i+1} \end{cases} \quad (2)$$

where $\Delta R_{i,i+1}$, $\Delta v_{i,i+1}$ and $\Delta p_{i,i+1}$ denote pre-integrated rotation, velocity, and position measurements.

The visual residual formulation we applied is (3), where $\prod$ is the camera projection function, $u_{ij}$ is the feature coordinate in an image plane, $x_j$ is the 3D landmark position in the local world frame, $\oplus$ is the transformation operation of the SE(3) group over $R^3$ elements, and $T_{CB}$ is the rigid extrinsic between IMU and camera. Then, the visual-inertial odometry can be posed as a keyframe-based nonlinear optimization problem, and the visual-inertial optimization problem can be written as (4), where $r_{I_{i,i+1}}$ is the inertial residual, $r_{ij}$ is the visual residual, and $\rho_{Hub}$ is a robust Huber kernel for decreasing the impact of incorrect matchings.

$$r_{ij} = u_{ij} - \Pi(T_{CB}T_i^{-1} \oplus x_j) \quad (3)$$

$$\min_{\overline{S}_k, \mathcal{X}} \left\{ \sum_{i=1}^{k} \left\| r_{\mathcal{I}_{i-1,i}} \right\|_{\Sigma_{\mathcal{I}_{i,i+1}}}^2 + \sum_{j=0}^{l-1} \sum_{i \in \mathcal{K}^j} \rho_{Hub}\left( \left\| r_{ij} \right\|_{\Sigma_{ij}} \right) \right\} \quad (4)$$

We formulate a novel GPS residual shown in Fig. 3 to correct the error of visual-inertial estimation. This residual is composed of the GPS data, IMU preintegration, IMU pose, IMU velocity, accelerometer bias, and gyroscope bias. This residual has relative constraint among IMU's states, bias, and preintegration at time $t_i$. To jointly minimize these errors and residuals, we propose an optimization-based state estimation method. Here, the non-linear optimization problem can be stated as:

$$\min_{\overline{S}_k, \mathcal{X}} \left\{ \sum_{i=1}^{k} \left\| r_{\mathcal{I}_{i-1,i}} \right\|_{\Sigma_{\mathcal{I}_{i,i+1}}}^2 + \sum_{j=0}^{l-1} \sum_{i \in \mathcal{K}^j} \rho_{Hub}\left( \left\| r_{ij} \right\|_{\Sigma_{ij}} \right) + \sum_{k=1}^{m} \left\| r_{g_k} \right\|_{\Sigma_{g_k}}^2 \right\} \quad (5)$$

where $r_{g_k}$ is the GPS residual.

## II. OUR APPROACH

### A. GPS-VIO Initialization

The entire system initialization is usually divided into two steps. Firstly, after starting the VIO system, the VIO initialization is performed to obtain an excellent initial inertial variable, and this step generally takes 15–30 s for the scale error to converge to 1% [19]. The second step is GPS-VIO initialization which is performed after the accurate VIO initialization [9, 18], so this step will also take a certain amount of time to collect the GPS measurements and corresponding positions in the VIO frame to align these two trajectories. To solve the time-consuming problem of GPS-VIO initialization, we propose a new GPS-VIO fast initialization approach, which is still divided into the above two parts, but our new method can synchronously collect and process the data required by the second step when the first step is performed, thus reducing the time that the second step originally needs to consume. The details of our method are as follows.

During the VIO initialization, we can obtain $p_{b_i}^{b_0}$ (the position of each IMU frame with respect to the first IMU frame) by stereo vision odometry and $P_{g_i}^{enu}$ (the GPS position of interpolating to the corresponding IMU frame) by GPS measurements, and transform $p_{b_i}^{b_0}$ into $p_{g_i}^{b_0}$ (the GPS positions related to the first IMU frame) via equation (6), where $p_g^b$ is the GPS-IMU extrinsic.

$$p_{g_i}^{b_0} = R_{b_i}^{b_0} p_g^b + p_{b_i}^{b_0} \quad (6)$$

$$\min \sum_{i=0}^{k} \left\| p_{g_i}^{enu} - R_{b_0}^{enu} p_{g_i}^{b_0} - p_{b_0}^{enu} \right\|^2 \quad (7)$$

Thus, we can obtain the trajectory $p_{g_i}^{b_0}$ through stereo vision odometry and the corresponding interpolated GPS trajectory in the ENU frame. Then, we align these two trajectories with calculating $R_{b_0}^{enu}$ and $p_{b_0}^{enu}$ by minimizing equation (7), a least-squares estimation problem that can be solved by the umeyama

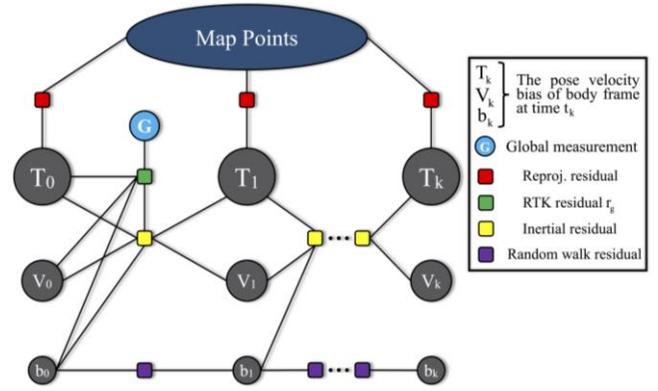

Fig. 3. Factor graph representation for tightly coupled optimization-based GPS-Visual-Inertial Odometry.

method.

During the GPS-VIO initialization, since we already get $R_{b_0}^{enu}$ and $p_{b_0}^{enu}$, we do not need to collect and align the GPS measurements and corresponding positions in the VIO frame. We can also get more accurate gravity direction $R_{b_0}^{w}$ after the scale error converges to 1%, meaning that scale and gravity directions are already accurately estimated. Finally, we can calculate $R_w^{enu}$ and $p_w^{enu}$ through the formulas (8).

$$\begin{cases} R_w^{enu} = R_{b_0}^{enu} R_{b_0}^{w\ T} \\ p_w^{enu} = -R_{b_0}^{enu} R_{b_0}^{w\ T} p_{b_0}^{w} + p_{b_0}^{enu} \end{cases} \quad (8)$$

Through this method, the initialization of VIO and GPS-VIO can be completed almost simultaneously, which improves the system efficiency compared with [9, 18].

### B. GPS Residual and Calibration

After finishing the initialization of the GPS-VIO system, the GPS measurements are integrated into the non-linear optimization framework. Then, to reduce the accumulated errors in the visual-inertial estimation and maintain the system's normal work in the case of GPS or visual failure, we propose a novel residual utilizing GPS measurements and IMU preintegration.

Firstly, we derive the formulas (9) to transform the GPS measurements $p_{g_k}^{enu}$ at time $t_k \in (t_i, t_{i+1})$ into the IMU frame $p_{b_{i+1}}^{w}$ at time $t_{i+1}$ for efficiently using IMU preintegration $\Delta p_{i+1}^{i}$ and $\Delta R_{i+1}^{i}$, where $t_d$ is the time offset between the GPS and IMU clock.

$$\begin{cases} p_{g_k}^{w} = R_w^{enu\ T}(p_{g_k}^{enu} - p_w^{enu}) \\ R_{b_k}^{w} = R_{b_i}^{w} \text{Exp}\left(\alpha \text{Log}\left(R_{b_i}^{w\ T} R_{b_{i+1}}^{w}\right)\right) \\ p_{b_k}^{w} = p_{g_k}^{w} - R_{b_k}^{w} p_g^{b} \\ p_{b_{i+1}}^{w} = \frac{1}{\alpha}(p_{b_k}^{w} - p_{b_i}^{w}) + p_{b_i}^{w} \\ \alpha = (t_k + t_d - t_i)/(t_{i+1} - t_i) \end{cases} \quad (9)$$

As shown in Fig. 3, our residual shown as the green square is composed of $\Delta R_{i+1}^{i}$, $\Delta p_{i+1}^{i}$, $T_{b_i}^{w}$, $v_{b_i}^{w}$, $b_{a_i}$ and $b_{w_i}$. We bring $p_{b_{i+1}}^{w}$ in (9) into formula (10) to construct the residual.

$$r_{g_k} = R_{b_i}^{w\ T}\left( p_{b_{i+1}}^{w} - p_{b_i}^{w} - v_{b_i}^{w} \Delta t_i + \frac{1}{2} g^w \Delta t_i^2 \right) - \Delta p_{i+1}^{i} \quad (10)$$

Due to space reasons, the formula (11) is not fully expanded. The GPS residual $r_{g_k}$ at timestamp $t_k$ is:

$$r_{g_k} = R_{b_i}^{wT}\left(\frac{1}{\alpha}(p_{g_k}^w - R_{b_k}^w p_g^b - p_{b_i}^w) - v_{b_i}^w \Delta t_i + \frac{1}{2}g^w \Delta t_i^2\right) - \Delta p_{i+1}^i \quad (11)$$

where $\Delta p_{i+1}^i$ and $\Delta R_{i+1}^i$ are separately the position and rotation preintegration using inertial measurements in the time interval $[t_i, t_{i+1}]$.

Since GPS-IMU extrinsic and time offset has a significantly impact on the GPS residual, we add the two parameters to the optimization states to enable online calibration of them. The GPS residual Jacobian of the GPS-IMU extrinsic and time offset is (12) and (13), respectively.

$$\frac{\partial r'_{g_k}}{\partial p_g^b} = -\frac{1}{\alpha}\text{Exp}\left(\alpha \text{Log}\left(R_{b_i}^{wT} R_{b_{i+1}}^{w}\right)\right) \quad (12)$$

$$\frac{\partial r'_{g_k}}{\partial t_d} = -\frac{t_{i+1} - t_i}{(t_k + t_d - t_i)^2} R_{b_i}^{wT}(p_{g_k}^w - R_{b_k}^w p_g^b - p_{b_i}^w) \\ + \frac{1}{(t_k + t_d - t_i)} R_{b_i}^{wT} R_{b_k}^w \lfloor p_g^b \times \rfloor J_l(\alpha \theta_i^{i+1}) \theta_i^{i+1} \quad (13)$$

where $[\times]$ is the skew symmetric matrix, $J_l$ is the left Jacobian of SO(3) [20], and $\theta_i^{i+1} = \log(R_{b_{i+1}}^{w\top} R_{b_i}^w)$.

After GPS-IMU extrinsic is added to the optimization states, the total states of GPS-Visual-Inertial Odometry are to be estimated by (14).

$$\mathcal{S}_i \doteq \left\{T_{b_i}^w, v_{b_i}^w, b_{a_i}, b_{w_i}, p_g^b, t_d\right\} \quad (14)$$

## III. EXPERIMENT

We perform three experiments to evaluate our proposed method. Firstly, we compare our proposed algorithm with a tightly coupled approach [18] on public EuRoC datasets [21]. EuRoC datasets are collected in the industrial machine hall and Vicon room providing global position measurement with millimeter-level precision from the motion capture system. To simulate real noisy GPS data, we add Gaussian noise to the global position measurements. Besides, to quantitatively evaluate the proposed approach against another optimization-based VIO-GPS pipeline, we assume to add the same magnitude of the noise. In the second experiment, we compare our proposed algorithm with the loosely coupled approach on common KAIST datasets [22]. In addition, compared with other tightly coupled approaches, our system is the first to be tested on the KAIST dataset. And we test the performance of our system when losing vision or GPS. Finally, we test our system to evaluate the performance and robustness in the outdoor scenes using our own MAV. We further test the performance and convergence of our online calibration method. As shown in Fig. 4, we use the UAV to collect image and IMU data. The global position measurement and the ground truth are obtained by RTK with centimeter-level accuracy.

So far ORB-SLAM3 is an excellent open-source algorithm with the highest accuracy, and our system is extended based on it. VINS-Fusion is a loosely coupled pose-graph optimization method on a sliding window. [18] tightly couples the GPS information in an optimized framework, and since this research is not open source, we can only use the same common EuRoC dataset to roughly evaluate the accuracy. In this section, we compare our system with ORB-SLAM3, VINS-Fusion and the tightly coupled approach on different datasets respectively. All experiments are performed on a laptop with 8GB of RAM and a 1.7GHz×8 Intel Core i5-8350U CPU.

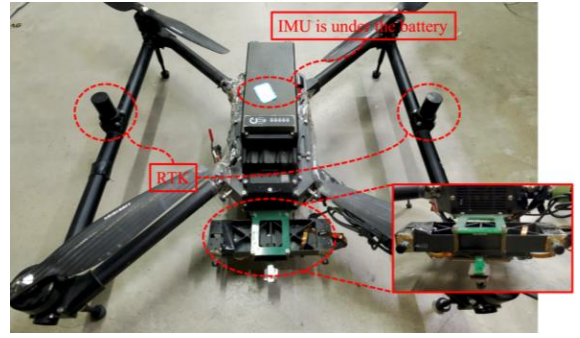

Fig. 4. The UAV with a stereo camera, hardware-synchronized IMU, and hardware-synchronized RTK engine.

### A. EuRoC datasets

EuRoC datasets are recorded on-board a MAV, using a front-down looking stereo camera (MT9V034 global shutter, 20Hz), hardware-synchronized with an IMU (ADIS16448, 200Hz). Meanwhile, a Leica Nova MS50 laser tracker or a Vicon motion capture system provides the ground truth states. EuRoC datasets contain two types of data. The first batch of datasets contain 5 sequences collected in the industrial machine hall. The second batch of datasets contains 6 sequences recorded in the Vicon room. The content of the datasets includes slow flights in an unstructured and cluttered room and fast flights in a large industrial machine hall, which renders the dataset challenging and representative to process. We add gaussian noise to the global position measurements in order to simulate real noisy GPS data. Gaussian noise is defined as $n_{mc} \sim \mathcal{N}(0, \sigma_{mc}^2 I)$, $\sigma_{mc} = 20\text{cm}$.

As usual in this field, we evaluate the proposed method in terms of trajectory estimation accuracy. We use the trajectory evaluation tool EVO to calculate and evaluate the trajectory estimation accuracy, and we measure accuracy with RMS ATE [23]. Table I compares the performance of our system using GPS-Stereo-IMU sensor configuration with the state-of-the-art algorithm ORB-SLAM3, loosely-coupled method VINS-Fusion and tightly coupled method [18]. Our results are the median after 10 executions. It is worth mentioning that we take the result of n = 1 in [18], because we only fuse one GPS data. As shown in the table, through tightly-coupled fusing GPS and visual-inertial odometry, our system achieves more accurate results than ORB-SLAM3 and VINS-Fusion. In addition, the accuracy of our system is also better than [18]. Compared to ORB-SLAM3, the mean ATE of EuRoC sequences reduced by more than 40%, with the highest decrease of 60% in sequence MH05. The mean ATE for all the EuRoC sequences is equal to 0.021 m. In the difficult machine hall sequences MH05, the illumination change is more than in other sequences. There is a dark scene in the middle of the sequence MH05, and accumulated error in the visual-inertial estimation becomes larger at this time. The result shows that the accumulated error can be reduced by fusing information from GPS and VIO. Compared to VINS-Fusion, the mean ATE of our system in EuRoC sequences is one-fifth of that of VINS-Fusion. As shown in Fig. 5. The trajectory of VINS-

TABLE I. ATE[M] COMPARISON IN THE EUROC DATASETS

| Sequence | MH01 | MH02 | MH03 | MH04 | MH05 | V101 | V102 | V103 | V201 | V202 | V203 | Avg |
|---|---|---|---|---|---|---|---|---|---|---|---|---|
| **ORB-SLAM3** | 0.036 | 0.037 | 0.032 | 0.057 | 0.083 | 0.037 | 0.016 | 0.027 | 0.029 | 0.014 | 0.026 | 0.035 |
| **VINS-Fusion** | 0.075 | 0.122 | 0.066 | 0.102 | 0.135 | 0.083 | 0.096 | 0.114 | 0.066 | 0.144 | 0.201 | 0.110 |
| **Tightly- coupled (Giovanni Cioff)** | 0.031 | 0.036 | 0.048 | 0.068 | 0.056 | 0.041 | 0.048 | 0.068 | 0.038 | 0.046 | 0.098 | 0.052 |
| **Tightly-coupled (ours)** | 0.023 | 0.026 | 0.024 | 0.029 | 0.031 | 0.019 | 0.012 | 0.021 | 0.019 | 0.014 | 0.017 | 0.021 |

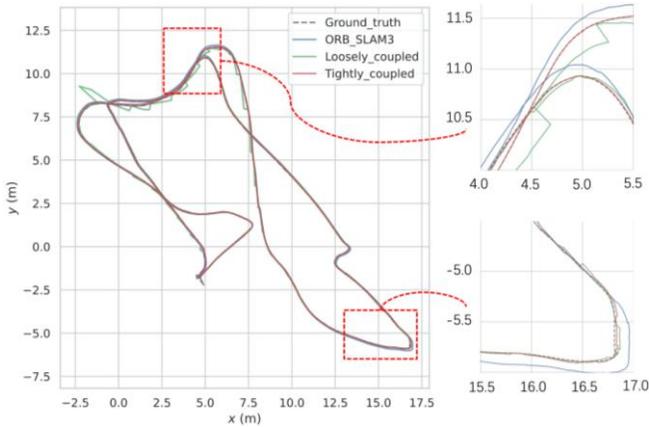

Fig. 5. Trajectory top-view of the EuRoC sequence MH05. The trajectories of Ground truth, ORB-SLAM3, loosely coupled method VINS-Fusion, and our tightly coupled methods are depicted. The two zoomed-in sections show that our tightly-coupled method reduces the ORB-SLAM3's accumulated error.

Fusion deviates greatly from the ground truth at the beginning. Since the transformation between the reference frames of between GPS and the VIO has not converged to the correct value at the beginning, the trajectory of VINS-Fusion will jump as the transformation changes. Compared to the tightly coupled method [18], the mean ATE of our system in EuRoC sequences is half of that. The difference in improvement relative to that method is likely due to two reasons. The first is that we estimated the extrinsic parameters and time offset online, and the second is we used different front-end and framework, which we use the stereo ORBSLAM3 front-end, and [18] uses the monocular SVO front-end. In general, our system trajectory is significantly better than ORB-SLAM3, the loosely coupled method VINS-Fusion and the tightly coupled approach.

### B. KAIST datasets

We further evaluate our system in a KAIST dataset, *urban26*, which is collected in urban area with 4 km long trajectory. KAIST datasets are recorded on a car, using a front-down looking stereo camera (Pointgrey Flea3, 10Hz), an IMU (Xsense MTi-300, 100Hz), VRS GPS (SOKKIA GRX2) and other sensors. Meanwhile, the ground-truth is the trajectory of the vehicle generated by various algorithms at the rate of 100 Hz.

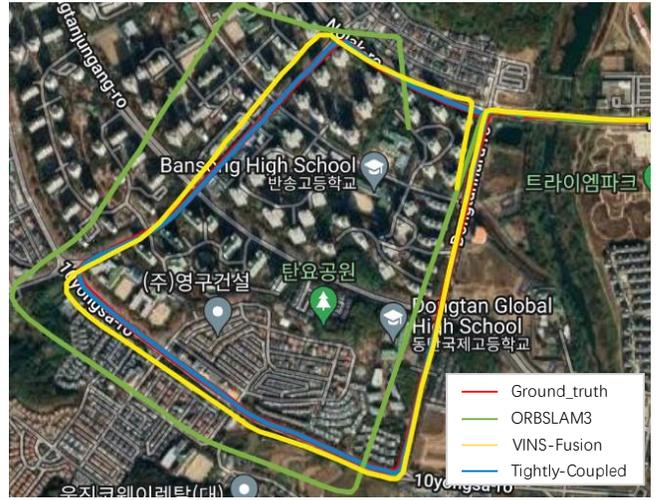

Fig. 6. The red line is the ground truth, brown is ORBSLAM3, green is VINS-Fusion, blue is our tightly coupled system.

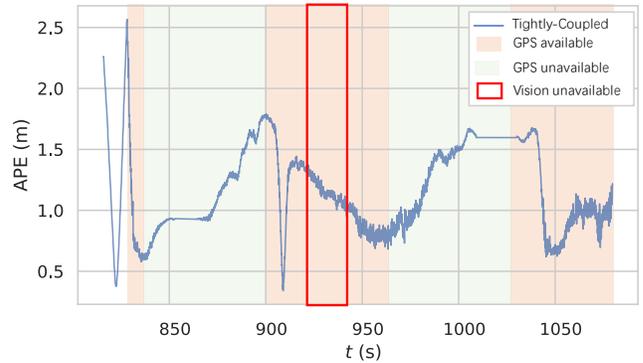

Fig. 7. The position error of our system in the case of visual or GPS failure.

We compare our result with ORB-SLAM3 and VINS-Fusion. Our results are the median after 10 executions. Fig. 6 shows the result of the experiment. The RMSE of our system, ORBSLAM3 and VINS-Fusion compared with the ground-truth is 1.23m, 115.56m, 24.84m respectively. Fig. 6 shows our system trajectory is significantly better than ORB-SLAM3 and VINS-Fusion. It is worth mentioning that the performance of VINS-Fusion on the EuRoC dataset and KAIST dataset is very different, because the loosely coupled method corrects the trajectory drift of VIO over long-term runs.

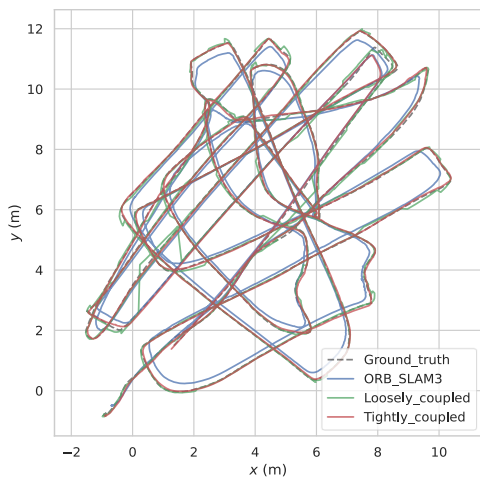

Fig. 8. Trajectory top-view in flight sequence 1. The trajectories of Ground truth, ORB-SLAM3, loosely coupled method VINS-Fusion, and our tightly coupled methods are depicted.

In addition, we test whether our system can work normally in the case of visual or GPS failure. As shown in Fig. 7, the white background is our fast initialization process, which took about 15 seconds. It is clear that the position estimation error of our tightly coupled system drops quickly when GPS measurements are available. When GPS is unavailable, the position error begins to gradually increase. In addition, we blur the image to test whether our system can work normally in the case of visual failure in the red box. The result shows the GPS residual can correct the IMU bias, our method can maintain the system's regular work in the case of visual failure.

*C. Outdoor experiments*

In the outdoor experiment, we choose the outdoor basketball court as the experiment area. As shown in Fig. 4, the UAV contains a stereo camera (20Hz), hardware-synchronized IMU (125Hz), and hardware-synchronized RTK engine (5Hz). RTK engine provides the receiver's location at an accuracy of 5cm in the open area. The datasets are recorded on-board the UAV in the basketball court around and around. We collect two flight sequences, and the flight distances of the two flight sequences are 210 m and 179 m respectively.

We compare our result with ORB-SLAM3 and VINS-Fusion. Our results are the median after 10 executions. The RMSE of our system, ORBSLAM3 and VINS-Fusion compared with the RTK is 0.092m, 0.278m, 0.145m respectively in the first flight, and the RMSE is 0.092m, 0.278m, 0.145m respectively in the second flight. Compared with ORB-SLAM3, the mean ATE of EuRoC sequences reduced by more than 66%. By fusing GPS measurements and VIO, our system achieves more accurate results than ORB-SLAM3 in the outdoor experiment. The position error of our system relative to the loosely-coupled method VINS-Fusion is also reduced by 36%. Fig. 8 shows Our system trajectory is significantly better than the loosely coupled method VINS-Fusion and ORB-SLAM3.

In addition, we test the performance of our online GPS-IMU extrinsic calibration method. Since we cannot know the accurate GPS-IMU extrinsic, the performance of online extrinsic calibration is evaluated by the robustness and convergence of multiple trials. To study the robustness of our online calibration method, we perform GPS-IMU extrinsic online calibration with different initial values. It can be clearly seen that the GPS-IMU extrinsic converges to near the hand-measured extrinsic value. This shows that the online extrinsic calibration can provide robust GPS-IMU extrinsic estimates.

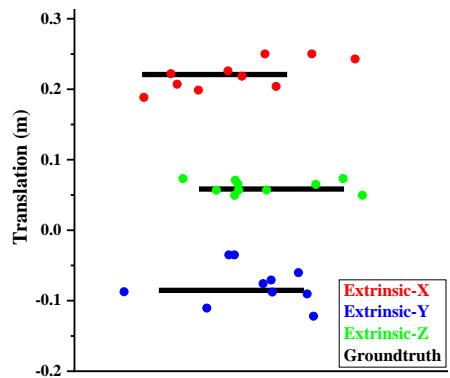

Fig. 9. Online calibration results of different initial GPS-IMU extrinsic in flight sequence 1.

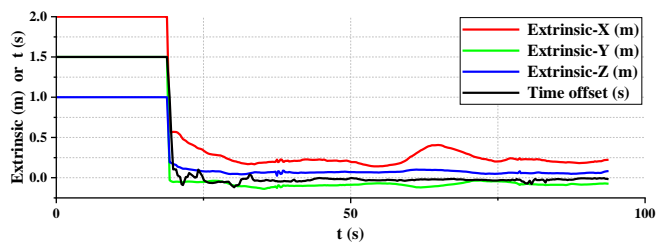

Fig. 10. the convergence of online GPS-IMU extrinsic calibration in flight sequence 1.

To study the convergence of our online calibration method, we perform GPS-IMU extrinsic and time offset calibration with poor initial values. The hand-measured extrinsic is $[0.21, -0.08, 0.06]^T$ meters and the initial value is $[2.00, 1.50, 1.00]^T$ meters. The GPS and IMU are hardware synchronized, and the initial value is 1.5s. Fig. 10 shows that the result converges to near hand-measured extrinsic and 0s respectively. Since the online initialization takes a little time, GPS-IMU extrinsic and time offset will have a straight line at the beginning. In general, this shows our online calibration method is stable.

IV. CONCLUSION

In this paper, we have developed a tightly coupled optimization-based GPS-Visual-Inertial odometry system that jointly minimizes visual reprojection errors, IMU errors, and GPS measurements errors formulated by GPS and IMU preintegration. To improve the robustness of the system, we propose an online calibration method for GPS-IMU extrinsic and time offset under a non-linear optimization framework. Then the performance and convergence of online GPS-IMU extrinsic and time offset calibration shows that our online calibration method is stable and robust. Meanwhile, we propose a fast initialization method, which improves the system efficiency. This system has been validated on EuRoC, KAIST and our outdoor datasets.

## REFERENCES

[1] Davison, "Real-time simultaneous localisation and mapping with a single camera," Proceedings Ninth IEEE International Conference on Computer Vision, 2003, pp. 1403-1410 vol.2.

[2] S. Sukkarieh, E. M. Nebot and H. F. Durrant-Whyte, "A high integrity IMU/GPS navigation loop for autonomous land vehicle applications," in IEEE Transactions on Robotics and Automation, vol. 15, no. 3, pp. 572-578, June 1999.

[3] Almagbile, Ali, Jinling Wang, and Weidong Ding. "Evaluating the performances of adaptive Kalman filter methods in GPS/INS integration." Journal of Global Positioning Systems 9.1 (2010): 33-40.

[4] L. M. Paz, P. PiniÉs, J. D. TardÓs and J. Neira, "Large-Scale 6-DOF SLAM With Stereo-in-Hand," in IEEE Transactions on Robotics, vol. 24, no. 5, pp. 946-957, Oct. 2008.

[5] R. Mur-Artal, J. M. M. Montiel, and J. D. Tard´os, "ORB-SLAM: a versatile and accurate monocular SLAM system," IEEE Transactions on Robotics, vol. 31, no. 5, pp. 1147–1163, 2015.

[6] R. Mur-Artal and J. D. Tard´os, "ORB-SLAM2: An open-source SLAM system for monocular, stereo, and RGB-D cameras," IEEE Transactions on Robotics, vol. 33, no. 5, pp. 1255–1262, 2017.

[7] Mur-Artal R, Tardós J D, "Visual-inertial monocular SLAM with map reuse," IEEE Robotics and Automation Letters, vol. 2, no. 2, pp. 796–803, 2017.

[8] S. Umeyama, "Least-squares estimation of transformation parameters between two point patterns," in IEEE Transactions on Pattern Analysis and Machine Intelligence, vol. 13, no. 4, pp. 376-380, April 1991.

[9] W. Lee, K. Eckenhoff, P. Geneva and G. Huang, "Intermittent GPS-aided VIO: Online Initialization and Calibration," 2020 IEEE International Conference on Robotics and Automation (ICRA), 2020, pp. 5724-5731.

[10] Shaozu Cao and Xiuyuan Lu and Shaojie Shen, "GVINS: Tightly Coupled GNSS-Visual-Inertial Fusion for Smooth and Consistent State Estimation," *arXiv preprint arXiv: 2103.07899*, 2021.

[11] Yongseok Lee, Jaemin Yoon, H. Yang, Changu Kim and D. Lee, "Camera-GPS-IMU sensor fusion for autonomous flying," 2016 Eighth International Conference on Ubiquitous and Future Networks (ICUFN), 2016, pp. 85-88.

[12] Xiaoying Kong, E. M. Nebot and H. Durrant-Whyte, "Development of a nonlinear psi-angle model for large misalignment errors and its application in INS alignment and calibration," Proceedings 1999 IEEE International Conference on Robotics and Automation (Cat. No.99CH36288C), 1999, pp. 1430-1435 vol.2.

[13] K. Hausman, J. Preiss, G. S. Sukhatme and S. Weiss, "Observability-Aware Trajectory Optimization for Self-Calibration With Application to UAVs," in IEEE Robotics and Automation Letters, vol. 2, no. 3, pp. 1770-1777, July 2017.

[14] Ramanandan, Arvind, Murali Chari, and Avdhut Joshi. "Systems and methods for using a global positioning system velocity in visual-inertial odometry." U.S. Patent No. 10,371,530. 6 Aug. 2019.

[15] Liu, Jinxu, Wei Gao, and Zhanyi Hu. "Optimization-Based Visual-Inertial SLAM Tightly Coupled with Raw GNSS Measurements." arXiv preprint arXiv:2010.11675, 2020.

[16] Qin, Tong, et al. "A general optimization-based framework for global pose estimation with multiple sensors." arXiv preprint arXiv:1901.03642, 2019.

[17] T. Qin, P. Li and S. Shen, "VINS-Mono: A Robust and Versatile Monocular Visual-Inertial State Estimator," in IEEE Transactions on Robotics, vol. 34, no. 4, pp. 1004-1020, Aug. 2018.

[18] G. Cioffi and D. Scaramuzza, "Tightly-coupled Fusion of Global Positional Measurements in Optimization-based Visual-Inertial Odometry," 2020 IEEE/RSJ International Conference on Intelligent Robots and Systems (IROS), 2020, pp. 5089-5095.

[19] C. Campos, R. Elvira, J. J. G. Rodríguez, J. M. M. Montiel and J. D. Tardós, "ORB-SLAM3: An Accurate Open-Source Library for Visual, Visual–Inertial, and Multimap SLAM," in IEEE Transactions on Robotics.

[20] G. Chirikjian, Stochastic Models, Information Theory, and Lie Groups, Volume 2: Analytic Methods and Modern Applications. Springer Science & Business Media, 2011, vol. 2.

[21] Burri, Michael, et al. "The EuRoC micro aerial vehicle datasets." The International Journal of Robotics Research 35.10 (2016): 1157-1163.

[22] Jeong J, Cho Y, Shin Y S, et al. Complex urban dataset with multi-level sensors from highly diverse urban environments[J]. The International Journal of Robotics Research, 2019, 38(6): 642-657.

[23] J. Sturm, N. Engelhard, F. Endres, W. Burgard and D. Cremers, "A benchmark for the evaluation of RGB-D SLAM systems," 2012 IEEE/RSJ International Conference on Intelligent Robots and Systems, 2012, pp. 573-580.